%% file: arxiv.tex
\documentclass{article}
\usepackage{iclr2027_conference,times}

\input{math_commands.tex}

\usepackage{hyperref}
\usepackage{url}
\usepackage{booktabs}
\usepackage{enumitem}

\usepackage{multirow}
\usepackage{diagbox}
\usepackage{xcolor}
\usepackage{colortbl}
\usepackage{pifont}
\usepackage{bbding}
\usepackage{fontawesome}
\usepackage{graphicx}
\usepackage{caption}

\definecolor{projecticon}{HTML}{2EA44F}

\usepackage[ruled]{algorithm2e}

\title{Talk Like You: Imitating How You Speak\\
in Real-Time Talking Head Generation}
\author{
\textbf{Baiqin Wang}$^{1,2}$ \quad
\textbf{Zhixing Ding}$^{1,2}$ \quad
\textbf{Jijie Li}$^{1,2}$ \quad
\textbf{Jiankuo Zhao}$^{1,2}$ \\
\textbf{Zhen Lei}$^{1,2,3,4}$ \quad
\textbf{Xiangyu Zhu}$^{1,2}$\thanks{Corresponding author.} \\
\normalfont
$^{1}$MAIS, Institute of Automation, Chinese Academy of Sciences \\
$^{2}$School of Artificial Intelligence, University of Chinese Academy of Sciences \\
$^{3}$CAIR, HKISI, Chinese Academy of Sciences \quad
$^{4}$SCSE, FIE, M.U.S.T. \\
\{wangbaiqin2024, zhen.lei, xiangyu.zhu\}@ia.ac.cn \\
\href{https://bq-wang0511.github.io/TalkLikeYou/}{\textcolor{projecticon}{\faGlobe}\ Project Page}\quad
\href{https://github.com/BQ-Wang0511/TalkLikeYou}{\raisebox{-0.18ex}{\includegraphics[height=1.1em]{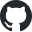}}\ GitHub}\quad
\href{https://huggingface.co/doubi-killer/TalkLikeYou}{\raisebox{-0.18ex}{\includegraphics[height=1.1em]{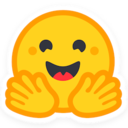}}\ Hugging Face}
}

\hypersetup{
  pdftitle={Talk Like You: Imitating How You Speak in Real-Time Talking Head Generation},
  pdfauthor={Baiqin Wang, Zhixing Ding, Jijie Li, Jiankuo Zhao, Zhen Lei, Xiangyu Zhu},
  pdfsubject={Real-Time Talking Head Generation},
  pdfkeywords={talking head generation, speaking habit imitation, flow matching}
}

\makeatletter
\renewcommand{\@maketitle}{
  \vbox{\hsize\textwidth
    \centering
    {\LARGE\scshape\@title\par}
    \vskip 0.22in
    {\normalsize\normalfont
      \begin{tabular}[t]{c}\@author\end{tabular}\par}
    \vskip 0.3in minus 0.1in
  }
}
\makeatother

\iclrfinalcopy
\begin{document}

\maketitle
\fancyhead{}
\renewcommand{\headrulewidth}{0pt}

\begin{abstract}
In daily life, each person exhibits unique speaking habits, leading to subtle yet consistent lip-shape variations even when pronouncing the same word. Although recent talking head generation methods have achieved impressive visual fidelity and lip synchronization, they largely overlook user-specific customization, especially the motion patterns that characterize individual speaking habits. These habits are difficult to model and capture, as their motion patterns are highly fine-grained and often similar across individuals. As a result, many approaches produce overly uniform facial motions and fail to capture diverse, person-specific articulation patterns. To address this, we propose TalkLikeYou, an efficient framework that imitates how a target person speaks in talking head generation. Our method models habit in motion-space and achieves real-time performance through Flow Matching with only one sampling step during inference. We further adopt a two-stage imitation learning strategy to capture subtle distinctions between habits, allowing users to specify a target habit through either a preset style from the dataset or a reference video. In addition, we introduce a new metric PLAD that projects mouth motions onto representative articulation axes to evaluate imitation accuracy and generation diversity. Extensive experiments demonstrate that TalkLikeYou generates high-quality talking heads in real-time and significantly improves speaking habit imitation compared with prior methods.
\end{abstract}

\section{Introduction}

\begin{figure*}[t] 
	\centering
	\includegraphics[width=\textwidth]{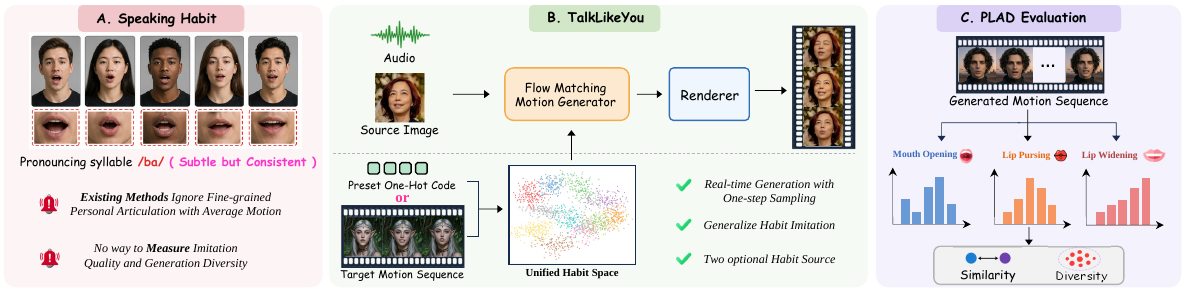}
	\vspace{-4mm}
	\caption{\textbf{Overview of TalkLikeYou.}
	Previous methods often generate uniform motions with limited control over speaking habits. We propose TalkLikeYou for efficient habit imitation in talking head generation, together with a new metric PLAD to evaluate habit similarity and diversity.}
	
	\label{fig:first}
	\vspace{-2mm}
\end{figure*}

Audio-driven talking head generation has attracted increasing attention due to its wide range of applications, including digital humans, voice assistants, and film-making. With rapid progress in this field~\cite{wang20243d,zhao2025stavatar}, customization has become an important direction for meeting user-specific needs, including imitating how a target person speaks. Such imitation is essential for building personalized virtual avatars that convey a recognizable identity and leave a vivid impression on viewers, as each person's speaking habits are reflected in subtle yet consistent articulatory behaviors in how the lips open, close, stretch, and transition during speech.  As shown in Fig.~\ref{fig:first}(a), even the same word can exhibit slight variations in articulation across individuals, such as the degree of mouth opening when saying ``b\textbf{u}s'', the lip width when pronouncing ``l\textbf{ea}ve'', or the pursing of the lips when saying ``f\textbf{oo}d''. These diverse habitual motion patterns are important parts of how we perceive identity and personality in daily communication~\cite{dobs2016identity}.

Yet imitating a person's speaking habits remains challenging, as they involve fine-grained lip motion patterns that are difficult to \textbf{model}. Previous methods~\cite{ji2025sonic,wang2025pc,cui2025hallo2,ki2025float} have made significant progress in generation quality, but they primarily focus on synchronizing lip movements with audio while overlooking subtle motion differences among individuals. They rely on image-space diffusion and typically learn average motion patterns from the dataset without explicitly distinguishing between speakers, resulting in limited diversity and personalization in lip motion. They also suffer from limited efficiency, as they require multiple sampling steps in high-dimensional spaces. Moreover, the differences between habitual motion patterns are often subtle, making them difficult for models to \textbf{capture}. Although some methods~\cite{ma2023styletalk,ye2024mimictalk} explore person-specific motion generation, they face limitations in imitation quality and motion coherence or rely on separate training for each individual. In addition, existing metrics are not designed to \textbf{measure} speaking-habit similarity and diversity. As a result, habit-aware talking-head generation remains underexplored.

To address these issues, we propose TalkLikeYou, an efficient talking head generation framework designed to imitate and control individual speaking habits, as shown in Fig.~\ref{fig:first}(b). To \textbf{model} speaking habits, our framework performs generation in motion space, providing better controllability, temporal coherence, and efficiency than image-space diffusion approaches~\cite{tian2024emo,ji2025sonic}.  Specifically, we first generate expressive head motions conditioned on target habitual motion and then render them into the final talking-head video. By incorporating Flow Matching~\cite{lipman2022flow} for motion synthesis, our method generates motions with a single sampling step during inference, enabling real-time performance. 

To \textbf{capture} fine-grained speaking-habit features for faithful imitation, we adopt a two-stage imitation learning strategy and establish a unified habit space using reference motion sequences and preset habit codes from the dataset. In the first stage, we initially train with only one-hot habit codes, as their explicit distinctions make habit differences easier for the model to learn. After this warm-up, we align reference-based habit features with their corresponding one-hot embeddings, then train on a large-scale video dataset in the second stage to improve generalization and diversity. This also allows our model to handle two scenarios: users can specify the target habit through a reference video or select a preset habit when no reference is available. It enables diverse and customizable generation while maintaining faithful speaking-habit imitation and accurate lip synchronization.

To \textbf{measure} the quality of speaking habit imitation and generation diversity, we introduce a new metric Projected Lip-Articulation Distribution (PLAD) as shown in Fig.~\ref{fig:first}(c). Since speaking habits are reflected in subtle variations in lip articulation, we project motion sequences onto three representative articulation axes and characterize the resulting distributions using histograms. As these distributions become stable for sufficiently long sequences, we compute  similarity and diversity scores from histograms to quantitatively evaluate speaking-habit imitation and generation diversity.

In summary, our contributions are as follows:

\begin{itemize}[leftmargin=20pt]
	\item We propose TalkLikeYou, a real-time talking-head generation framework capable of imitating specific speaking habits. By applying Flow Matching in motion space, our approach enables one-step motion generation with high coherence and efficiency.
	\item To improve imitation quality, we design a unified habit space and adopt a two-stage imitation learning strategy to improve imitation generalization and diversity.
	\item We develop a new metric PLAD to evaluate imitation quality by projecting motion sequences onto lip-articulation axes and measuring similarity and diversity across the histograms.
	\item Extensive experiments demonstrate that our method achieves state-of-the-art performance in talking-head generation, with superior lip synchronization and speaking-habit imitation.
\end{itemize}

\section{Related Work}

\subsection{Audio-Driven Talking Head Generation}
Talking head generation aims to synthesize realistic facial animations with lip movements synchronized to input audio. Early methods~\cite{wang2025talk,zhang2024musetalk} mainly focus on lower-face synthesis for video dubbing, such as Wav2Lip~\cite{prajwal2020lip}, which introduces a lip-sync expert network to improve synchronization. Recent approaches utilize modern generative models for better quality. For example, Sonic~\cite{ji2025sonic} and FLOAT~\cite{ki2025float} adopt DiT~\cite{peebles2023scalable} as their backbone, while MultiTalk~\cite{kong2025let} is built upon the video generation model WAN~\cite{wan2025wan}. However, these approaches often rely on heavy architectures, leading to high computational cost and long inference time. Moreover, they provide limited controllability for fine-grained motion patterns such as speaking habits.

In parallel, two-stage pipelines have gained attention. They first predict intermediate facial motion representations from audio, such as 3DMM parameters~\cite{FLAME:SiggraphAsia2017} or facial landmarks, and then render lip-synced videos conditioned on them. Among these representations, implicit keypoints show strong expressiveness: face-vid2vid~\cite{wang2021one} disentangles them into expression and pose, while LivePortrait~\cite{guo2024liveportrait} improves controllability with landmark-bounded keypoints. Several methods follow this paradigm: EAT~\cite{gan2023efficient} suffers from limited visual quality, while VASA~\cite{xu2024vasa} and Ditto~\cite{li2025ditto} apply diffusion in motion space but require multiple sampling steps and tend to generate uniform motions without modeling speaking habits. PC-Talk~\cite{wang2025pc} provides stronger control and lip synchronization, but its backbone limits generalization. Overall, two-stage methods remain promising, yet still leave room for improvement in efficiency and controllable habit-aware generation.

\subsection{Customized Talking Head Generation}

Customization has attracted increasing attention in talking-head generation. Several methods focus on emotion control~\cite{gan2023efficient,tan2024edtalk,wang2025pc}, generating emotional talking-face videos where the emotion source is derived from labels or other inputs such as text and audio. Other works incorporate image editing into this task, including hair transfer~\cite{feng2025faceedittalker,li2026devil} and face swapping~\cite{guan2024resyncer} for appearance modification. Some methods~\cite{zhu2025infp,wang2026towards} further focus on multi-person conversation to improve interaction. However, these works mainly address high-level customization, such as emotion, appearance, or interaction, rather than fine-grained motion control.

Meanwhile, several methods explore more detailed customization, such as imitating person-specific motion patterns, but they still face limitations. StyleTalk~\cite{ma2023styletalk} and DreamTalk~\cite{ma2023dreamtalk} perform conditioned synthesis with specific motion patterns, yet their control mainly targets emotion-related dynamics rather than habit-related variations in lip shape and articulation. MimicTalk~\cite{ye2024mimictalk} uses an autoregressive paradigm to reproduce person-specific habits, but it is person-dependent and requires training or adaptation for each individual, limiting generalization to unseen identities. PC-Talk~\cite{wang2025pc} can generate style-conditioned results, but may suffer from subtle temporal jitter and unstable motion. Although some 3D facial animation methods~\cite{fan2022faceformer,thambiraja2023imitator,zhou2025meta} also investigate this direction, their generated 3D motions cannot be directly applied to 2D talking-head generation. Overall, imitating how a target person speaks still leaves substantial room for improvement.

\section{Methods}

\subsection{Overall Framework}

\begin{figure*}[t] 
	\centering
	\includegraphics[width=\textwidth]{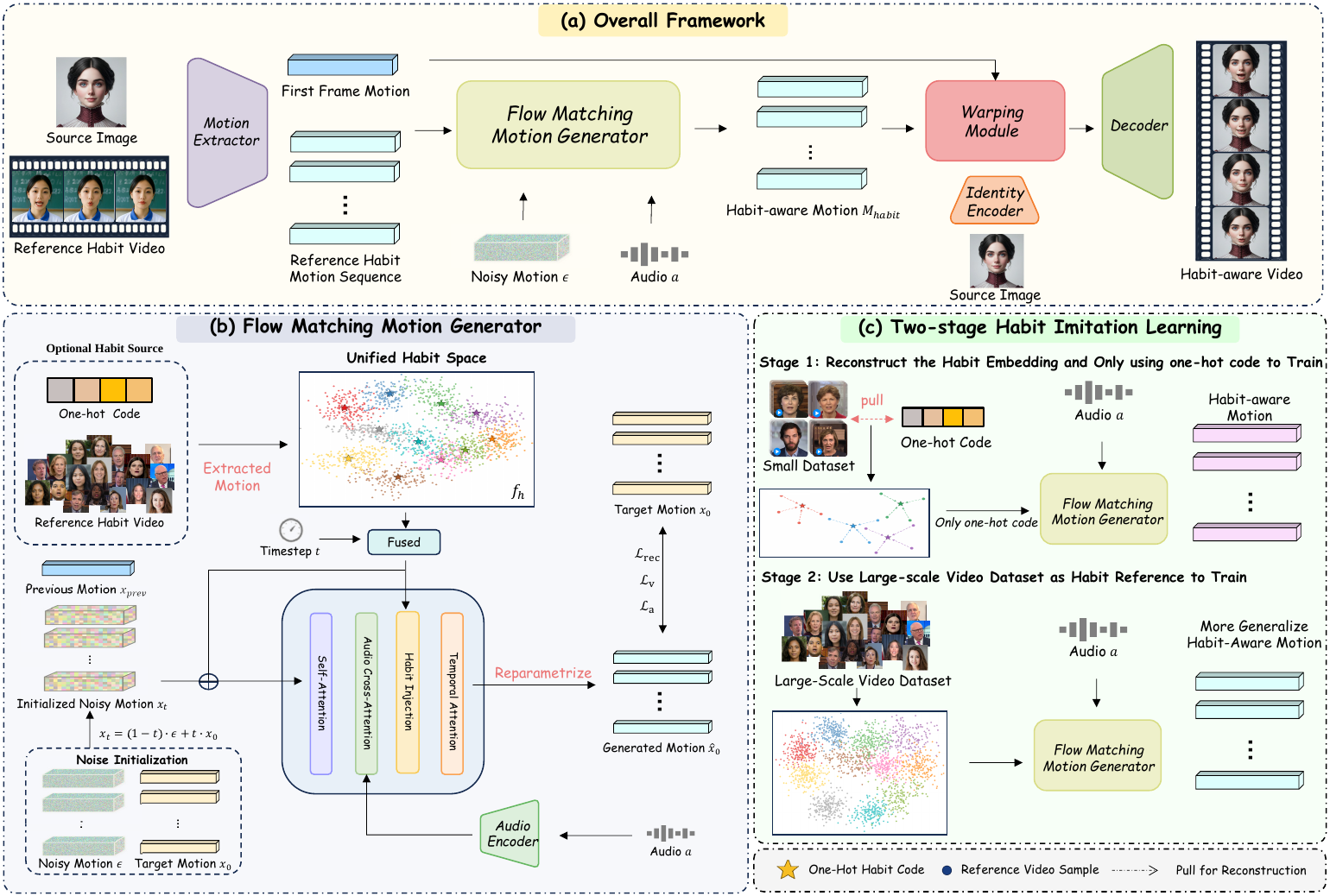}
	\vspace{-5mm}
	\caption{\textbf{Framework of TalkLikeYou.}
		Our method first generates habit-aware motion using the Flow Matching Motion Generator with optional habit sources through one-step sampling, and then renders them into a video. We further adopt a two-stage habit imitation learning strategy for more generalized and stable habit imitation. In the unified habit space, $\star$ denotes the one-hot habit code and $\bullet$ denotes reference-video samples. During the first training stage, we use an alignment loss between reference video features and their corresponding one-hot habit codes.}
	\label{fig:framework}
	\vspace{-3mm}
\end{figure*}

To imitate how a target person speaks, our framework first generates habit-aware talking motions and then renders them into the final video, as illustrated in Fig.~\ref{fig:framework}(a). Specifically, we use a motion extractor to obtain implicit keypoint-based motion representations from the source image and the reference video. Following the rendering pipeline~\cite{guo2024liveportrait}, the extracted motion is decomposed into canonical keypoints $K_c$, expression deformation $\delta$, and pose parameters, including rotation $R$, translation $t$, and scale $s$. The original keypoints $K_{ori}$ are formulated as:
\begin{equation}
	\begin{aligned}
		K_{ori} = s \cdot (K_c \cdot R + \delta) + t.
	\end{aligned}
	\label{eq:habit_transform}
\end{equation}

After obtaining the motion representations, we feed them into a Flow Matching-based motion generator to produce habit-aware motion. The generator is conditioned on habit feature $f_h$, the audio feature $f_{\text{aud}}$, and the source motion \(M_{\text{src}}\) used for initialization. It predicts the habitual motion $M_{\text{habit}}$, which is decomposed into the habit-aware rotation $R_h$ and expression deformation $\delta_h$:
\begin{equation}
	\begin{aligned}
		M_{\text{habit}} = G_{\theta}(f_h, f_{\text{aud}}, M_{\text{src}}),
		\quad
		M_{\text{habit}} \Rightarrow \{R_h, \delta_h\}.
	\end{aligned}
	\label{eq:habit_motion}
\end{equation}

We then apply the generated motion components to transform the original keypoints into habit-aware driven keypoints $K_h$:
\begin{equation}
	\begin{aligned}
		K_h = s \cdot (K_c \cdot R_h + \delta_h) + t.
	\end{aligned}
	\label{eq:habit_kh}
\end{equation}

Finally, we render the talking head video from the source image and the generated habit-aware motion. We first encode the source image into an appearance embedding $f_a$ using an identity encoder. A warping decoder then fuses $f_a$ with the original keypoints $K_{ori}$ and the habit-aware driven keypoints $K_h$ to estimate the dense warping field and synthesize the final rendered frame \(I_\text{res}\):
\begin{equation}
	\begin{aligned}
		I_\text{res} = \text{Decoder}\!\left(\text{Warp}(f_a, K_{ori}, K_h)\right).
	\end{aligned}
	\label{eq:habit_warp}
\end{equation}

Since implicit keypoints are semantically associated with facial components, we can selectively edit the lip-related keypoints from the input video to achieve video dubbing. During the training of the Flow Matching Motion Generator, all other parts of the rendering pipeline remain fixed, as they only serve as intermediate representations for motion generation and final video rendering.

\subsection{Flow Matching Motion Generator}

\noindent\textbf{Model Architecture.}
Our motion generator aims to produce talking motions that are synchronized with the input audio while reflecting the target speaking habit in the lip region. To this end, we employ a DiT~\cite{peebles2023scalable} composed of four progressive layers as shown in Fig.~\ref{fig:framework}(b). We first use self-attention to extract intrinsic motion representations, then apply audio cross-attention to incorporate audio features for synchronization. Next, we project either the input habit sequence or the preset one-hot habit code from the training dataset into a unified habit space, and fuse it through a habit-injection layer to enable target-habit imitation. Finally, a temporal attention layer refines sequence-level dynamics and improves temporal consistency. This design enables coherent talking motion generation while preserving the target speaking habit.

\noindent\textbf{Flow Matching Training.}
For efficient and high-quality generation, we adopt Flow Matching as our training objective. Conventional Flow Matching~\cite{lipman2022flow} learns a velocity field \(v_t\) to transport noisy samples toward the data distribution, enabling faster convergence than denoising-based samplers such as DDIM~\cite{song2020denoising} due to its simple linear trajectory. Although this formulation has shown strong performance in generative models such as FLUX~\cite{batifol2025flux}, we observe that directly predicting the velocity field can introduce noticeable jitter in generated motions. To address this, we reparameterize the model to predict the clean motion \(x_0\) at each timestep, which improves motion coherence and enables one-step sampling during inference. Specifically, at each training step \(t\), we sample noise motion \(\epsilon \sim \mathcal{N}(0, I)\) and construct the noisy motion as:
\begin{equation}
x_t = (1-t)\epsilon + t x_0, \quad t \sim \mathcal{U}(0,1).
\end{equation}

The previous motion \(x_{\mathrm{prev}}\) is obtained from source \(M_{\text{src}}\). Given the noisy motion \(x_t\), audio feature \(f_{\text{aud}}\), habit feature \(f_h\),  and timestep \(t\), our model predicts the clean motion:
\begin{equation}
M_{\text{src}} \Rightarrow x_{\mathrm{prev}}, \quad   
\hat{x}_0 = G_\theta(x_t, f_{\text{aud}}, f_h, x_{\mathrm{prev}}, t), \quad 
\hat{x}_0 \Rightarrow M_{\text{habit}}. 
\end{equation}

During training, we first apply an \(L_2\) reconstruction loss \(\mathcal{L}_{\mathrm{rec}}\) between the predicted motion \(\hat{x}_0\) and the ground-truth motion \(x_0\). Additionally, we impose a first-frame loss  \(\mathcal{L}_{f}\) that aligns the first generated frame with the last frame of the previous motion to ensure a seamless transition between consecutive context windows. The losses \(\mathcal{L}_{\mathrm{rec}}\) and \(\mathcal{L}_{f}\) are calculated as:
\begin{equation}
\mathcal{L}_{\mathrm{rec}} = \| \hat{x}_0 - x_0 \|_2^2, \quad
\mathcal{L}_{f} =\| \hat{x}_{0}^{0} - x_{\mathrm{prev}} \|_2^2.
\end{equation}
To further enhance temporal consistency, we introduce first-order and second-order temporal derivative losses to constrain frame-to-frame motion dynamics and regularize temporal smoothness:
\begin{equation}
\begin{aligned}
	\mathcal{L}_{v}
	&=
	\sum_{i=2}
	\left\|
	\Delta \hat{x}_0^{i} - \Delta x_0^{i}
	\right\|_2^2,
	\quad
	\Delta x_0^{i} = x_0^{i} - x_0^{i-1}, \\
	\mathcal{L}_{a}
	&=
	\sum_{i=3}
	\left\|
	\Delta^2 \hat{x}_0^{i} - \Delta^2 x_0^{i}
	\right\|_2^2,
	\quad
	\Delta^2 x_0^{i} = x_0^{i} - 2x_0^{i-1} + x_0^{i-2}.
\end{aligned}
\end{equation}

where \(\hat{x}_{0}^{0}\) denotes the first frame of the predicted motion. This constraint helps extend generation to arbitrary-length sequences by maintaining continuity across context windows. The overall training objective is formulated as:
\begin{equation}
\mathcal{L}
=
\mathcal{L}_{\mathrm{rec}}
+
\lambda_v \mathcal{L}_{v}
+
\lambda_a \mathcal{L}_{a}
+
\mathcal{L}_{f}.
\end{equation}

With this motion-spaced Flow Matching paradigm, our model can efficiently generate temporally coherent talking motions with accurate lip synchronization and faithful speaking habit imitation.

\subsection{Two-stage Habit Imitation Learning Strategy}
Although our framework can generate motions with a target habit from a reference motion sequence, directly training the model from scratch with reference motion sequences leads to unsatisfactory habit imitation. This is because speaking habits are fine-grained lip region patterns that often appear as subtle pronunciation-dependent variations, making them difficult to extract from motion alone. In contrast, one-hot habit codes provide explicit and separable supervision, making habit distinctions easier to learn and naturally supporting preset habit control when no reference video is available. Based on this, we propose a two-stage habit imitation strategy, as shown in Fig.~\ref{fig:framework}(c), to improve habit imitation quality while preserving preset one-hot control.

In the first stage, we train the motion generator on a limited dataset using only one-hot habit codes defined by training dataset, where each identity is assigned a unique code to represent its distinct talking habit. This explicit categorization helps the model learn audio-lip synchronization while distinguishing habit differences, forming a foundational habit space with separable representations. Meanwhile, we train the habit extractor with reference videos from the same dataset by reconstructing the corresponding one-hot habit feature:
\begin{equation}
\begin{aligned}
	f_h^{\mathrm{oh}} &= E_{\mathrm{oh}}(c), \quad
	\hat{x}_0 = G_\theta(x_t, f_{\text{aud}}, f_h^{\mathrm{oh}}, x_{\mathrm{prev}}, t), \\
	\hat{f}_h^{\mathrm{ref}} &= E_{\mathrm{ref}}(M_{\mathrm{ref}}), \quad
	\mathcal{L}_{\mathrm{align}} =
	\left\| \hat{f}_h^{\mathrm{ref}} - f_h^{\mathrm{oh}} \right\|_2^2,
\end{aligned}
\end{equation}
where $c$ denotes the one-hot identity code. This encourages reference-based habit features to be projected into a unified habit space, preparing the model for the next stage.

In the second stage, we train the model with reference videos from a large-scale dataset. For each sample, we select two sequences from the same person, where one sequence serves as the target motion and the other serves as the reference motion:
\begin{equation}
\hat{x}_0 = G_\theta(x_t, f_{\text{aud}}, \hat{f}_h^{\mathrm{ref}}, x_{\mathrm{prev}}, t).
\end{equation}

With the structured habit space learned in the first stage, the model can better imitate reference-derived habit features and learn diverse motion patterns from large-scale data. To preserve compatibility between two conditioning options, we periodically reuse the one-hot codes from the first stage during training, once every ten steps. This strategy improves habit imitation quality while supporting both reference video and preset one-hot habit control.

\section{Projected Lip-Articulation Distribution}

\setlength{\textfloatsep}{6pt}
\begin{algorithm}[t]
	\SetAlgoNoLine
	\KwIn{Generated motion sequences $\{M_i\}_{i=1}^{N}$; lip-articulation projection axes $\mathcal{Q}$.}
	\KwOut{PLAD histograms $\{\mathcal{H}_i\}_{i=1}^{N}$, pairwise habit similarity scores, and diversity score $D_{\mathrm{div}}$.}
	\For{each motion sequence $M_i$}{
		Extract expression deformation sequence $\delta_i=\{\delta_{i,t}\}_{t=1}^{T}$\;
		\For{each articulation axis $q \in \mathcal{Q}$}{
			\For{each frame $t$}{
				Compute projection scalar
				$p_{i,t}^{q} =
				\frac{\langle \delta_{i,t}, q\rangle}
				{\langle q,q\rangle}$\;
			}
			Construct normalized histogram $H_i^q=\mathrm{Hist}(\{p_{i,t}^{q}\}_{t=1}^{T})$\;
		}
		Form PLAD descriptor $\mathcal{H}_i=\{H_i^q \mid q \in \mathcal{Q}\}$\;
	}
	\For{each pair $(M_i,M_j)$}{
		Compute histogram IoU $S_{\mathrm{IoU}}(M_i,M_j)$ and JS distance $D_{\mathrm{JS}}(M_i,M_j)$ using Eq.~\ref{eq:hist_similarity}\;
	}
	Compute group diversity $D_{\mathrm{div}}$ using Eq.~\ref{eq:diversity}\;
	\caption{Projected Lip-Articulation Distribution for Speaking-Habit Evaluation}
	\label{alg:plad_assessment}
\end{algorithm}

Existing talking head metrics mainly measure visual quality or lip synchronization, but are not designed to evaluate whether generated motion preserves a specific speaking habit. To evaluate speaking-habit imitation and generation diversity, we introduce Projected Lip-Articulation Distribution (PLAD). Since speaking habits are reflected in subtle and consistent lip articulations, PLAD projects motion sequences onto representative lip-articulation axes and characterizes the resulting distributions using histograms. We assume that a speaking habit is not determined by isolated frames, but by stable statistical tendencies over repeated articulatory movements. Although each frame is influenced by phonetic content, a person tends to show consistent distributional patterns over sufficiently long sequences along these representative axes such as mouth opening, lip stretching, or lip pursing. Thus, motions with the same habit should have similar PLAD histograms, while different habits should yield distinguishable distributions.

Specifically, as shown in Alg.~\ref{alg:plad_assessment}, for a generated motion sequence, we extract the lip-region expression deformation sequence $\delta=\{\delta_i\}_{i=1}^{T}$, where each $\delta_i$ denotes the implicit expression parameters at frame $i$. We define a set of lip-articulation projection axes $\mathcal{Q}=\{q_m\}_{m=1}^{M}$, where each axis corresponds to a representative articulation pattern. For each frame, we project the expression parameters onto each articulation axis and obtain a scalar projection:
\begin{equation}
	p_i^{q} =
	\frac{\left\langle \delta_i, q \right\rangle}
	{\left\langle q, q \right\rangle},
	\quad q \in \mathcal{Q}.
	\label{eq:plad_projection}
\end{equation}

In practice, we construct three representative axes for \textit{mouth opening}, \textit{lip widening}, and \textit{lip pursing}, corresponding to articulations in words such as ``b\textbf{u}s'', ``l\textbf{ea}ve'', and ``f\textbf{oo}d'', respectively. The projection sequence $P^q=\{p_i^q\}_{i=1}^{T}$ describes how strongly the motion follows the articulation axis $q$ over time. We then construct a normalized histogram from each projection sequence and represent the final PLAD descriptor as:
\begin{equation}
	H^q = \mathrm{Hist}(P^q), \quad
	\mathcal{H} = \{H^q \mid q \in \mathcal{Q}\}.
	\label{eq:plad_descriptor}
\end{equation}
\begin{figure*}[t] 
	\centering
	\includegraphics[width=\textwidth]{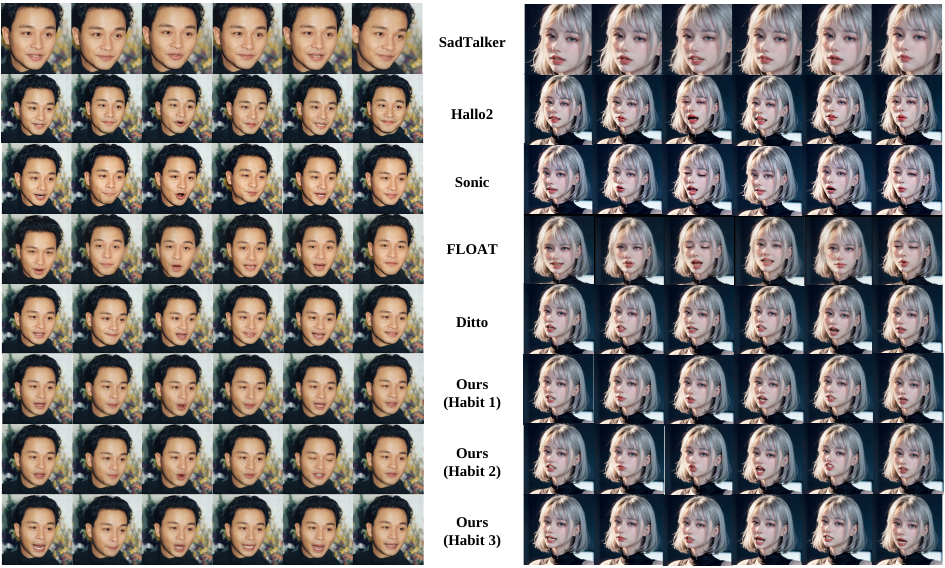}
	\vspace{-6mm}
	\caption{\textbf{Comparison Results.}
		We compare our method with advanced baselines under three selected speaking habits, showing better visual quality and lip synchronization. Please zoom in for details. Video results are provided in the supplementary material.  }
	\label{fig:comparison}
\end{figure*}
To compare two generated motions, we directly measure the similarity between their histograms. Given two motion sequences $A$ and $B$, we compute the histogram IoU and JS distance as:
\begin{equation}
	\begin{aligned}
		S_{\mathrm{IoU}}(A,B)=
		\frac{1}{|\mathcal{Q}|}
		\sum_{q \in \mathcal{Q}}
		\frac{\sum_b \min(H_A^q(b), H_B^q(b))}
		{\sum_b \max(H_A^q(b), H_B^q(b))}, \quad
		D_{\mathrm{JS}}(A,B)=
		\frac{1}{|\mathcal{Q}|}
		\sum_{q \in \mathcal{Q}}
		\mathrm{JS}(H_A^q \,\|\, H_B^q).
	\end{aligned}
	\label{eq:hist_similarity}
\end{equation}
where $b$ indexes histogram bins and $\mathrm{JS}(\cdot\|\cdot)$ denotes the Jensen-Shannon divergence. A higher histogram IoU and a lower JS distance indicate stronger habit consistency, while lower IoU and higher JS distance indicate better habit separability. Furthermore, to evaluate the diversity of speaking habits generated by a model, we compute the average pairwise JS distance over a group of generated samples. Given $N$ generated motion sequences, the diversity score is defined as:
\begin{equation}
	D_{\mathrm{div}}
	=
	\frac{2}{N(N-1)}
	\sum_{1 \le i < j \le N}
	D_{\mathrm{JS}}(M_i,M_j).
	\label{eq:diversity}
\end{equation}

The diversity score \(D_{\mathrm{div}}\) reflects whether a model can generate distinguishable speaking habits instead of collapsing to similar  distributions.

\section{Experiments}

\subsection{Experimental Settings}

\noindent\textbf{Implementation Details.}
We convert all videos to 25 fps and extract audio and motion representations before training. The Flow Matching Motion Generator is optimized with Adam~\cite{kingma2014adam} at a learning rate of \(1\times10^{-4}\), while other modules are fixed as they are initialized from LivePortrait~\cite{guo2024liveportrait} for its balance between efficiency and quality. The first and second stages are trained for 48 and 72 hours on an RTX 3090, respectively. We use HDTF~\cite{zhang2021flow} in the first training stage, which contains high-quality videos of 380 identities, each represented by a one-hot code as a preset habit option. The second stage is trained on CelebV-HQ~\cite{zhu2022celebv}, a large-scale in-the-wild talking head dataset. We adopt the Wav2Lip~\cite{prajwal2020lip} audio encoder for its audio-lip aligned features.  More details are provided in the supplementary.

\noindent\textbf{Baselines.}
We compare our method with five advanced talking head generation baselines, including SadTalker~\cite{zhang2023sadtalker}, Hallo2~\cite{cui2025hallo2}, Sonic~\cite{ji2025sonic}, FLOAT~\cite{ki2025float}, and Ditto~\cite{li2025ditto}.  For habit imitation, we choose StyleTalk~\cite{ma2023styletalk}, MimicTalk~\cite{ye2024mimictalk} and PC-Talk~\cite{wang2025pc}, as they are the only relevant open-source methods. Note that MimicTalk requires person-specific training before generating results.

\noindent\textbf{Metrics.}
We evaluate our method from several aspects. Lip synchronization is measured by Sync-C and Sync-D, which are computed using a pretrained SyncNet~\cite{chung2016out}. Image quality is evaluated with the reference-based FID~\cite{Seitzer2020FID} metric and the non-reference-based NIQE~\cite{mittal2012making} metric. We use FVD~\cite{unterthiner2019fvd} to measure temporal consistency and CSIM~\cite{deng2019arcface} for identity preservation. For habit imitation, we use the similarity and diversity metrics proposed in our PLAD.
\begin{table*}[t]
	\centering
	\caption{Quantitative comparisons with state-of-the-art methods. }
	\vspace{-3mm}
\resizebox{\textwidth}{!}{
	\begin{tabular}{c|cccccc|cccccc}
		\toprule
		\multicolumn{1}{c}{\multirow{2}[4]{*}{\textbf{Method}}} & \multicolumn{6}{c}{\textbf{HDTF}} & \multicolumn{6}{c}{\textbf{CelebV-HQ}}\\
		\cmidrule(lr){2-7}  \cmidrule(lr){8-13}  \multicolumn{1}{c}{} &
		\multicolumn{1}{c}{Sync-C$\uparrow$}& \multicolumn{1}{c}{Sync-D$\downarrow$}  & \multicolumn{1}{c}{FID$\downarrow$}  & \multicolumn{1}{c}{$\text{NIQE}\downarrow$} & \multicolumn{1}{c}{$\text{FVD}\downarrow$} & \multicolumn{1}{c}{$\text{CSIM}\uparrow$} & \multicolumn{1}{c}{Sync-C$\uparrow$} & \multicolumn{1}{c}{Sync-D$\downarrow$}  & \multicolumn{1}{c}{FID$\downarrow$} & \multicolumn{1}{c}{$\text{NIQE}\downarrow$} & \multicolumn{1}{c}{$\text{FVD}\downarrow$} & \multicolumn{1}{c}{$\text{CSIM}\uparrow$}
		\\
		\midrule
		
		SadTalk~\cite{zhang2023sadtalker} & 7.66 & 7.32 & 87.60 & 45.72 & 534.14 & 0.85 & 6.36 & 8.36 & 109.11 & 40.82 & 829.09 & 0.80 \\
		Hallo2~\cite{cui2025hallo2} & 7.33 & 7.97 & 25.41 & 13.02 & 272.57 & 0.83 & 6.32 & 8.73 & 63.95 & 13.13 & \textbf{688.03} & 0.75 \\
		Sonic~\cite{ji2025sonic} & \underline{8.62} & \textbf{6.70} & 23.42 & 12.91 & \underline{195.21} & 0.84 & \underline{7.68} & \underline{7.35} & 66.37 & 12.88 & 704.12 & 0.76 \\
		Float~\cite{ki2025float} & 7.04 & 8.16 & 84.70 & \underline{12.39} & 382.93 & 0.82 & 6.44 & 8.64 & 137.23 & \underline{11.89} & 854.89 & 0.70 \\
		Ditto~\cite{li2025ditto} & 5.23 & 9.94 & \underline{21.61} & 12.62 & 319.03 & \underline{0.87} & 5.80 & 9.19 & \underline{62.67} & 12.13 & 743.16 & \underline{0.82} \\
		\midrule
		\textbf{TalkLikeYou (Ours)} & \textbf{8.73} & \underline{6.82} & \textbf{7.71} & \textbf{12.24} & \textbf{112.94} & \textbf{0.94} & \textbf{7.97} & \textbf{7.25} & \textbf{61.51} & \textbf{11.79} & \underline{694.40} & \textbf{0.85} \\ 
		
		\bottomrule
		
	\end{tabular}
}
\label{tab:compare_all}
\end{table*}

\subsection{Evaluation on Lip-Synchronization}

\begin{figure}[t]
	\centering
	
	\begin{minipage}[t]{0.6\linewidth}
		\vspace{0pt}
		\centering
		\includegraphics[width=0.98\linewidth]
		{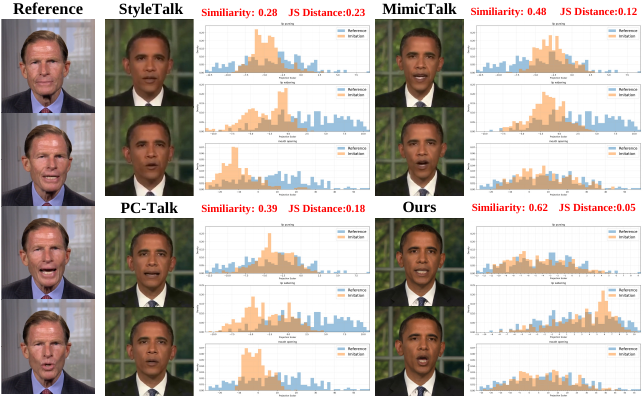}
		\vspace{-2mm}
		\captionof{figure}{Habit Imitation Comparison with related methods.}
		\label{fig:adaption}
	\end{minipage}
	\hfill
	\begin{minipage}[t]{0.37\linewidth}
		\vspace{0pt}
		\centering
		\scriptsize
		\vspace{3mm}
		\captionof{table}{Habit Diversity and Efficiency Comparison in each method.}
		\vspace{1mm}
		\label{tab:eff}
		
		\renewcommand{\arraystretch}{1.1}
		\resizebox{0.95\linewidth}{!}{
			\begin{tabular}{lcc}
				\toprule
				\textbf{Method}
				& \textbf{Diversity}
				& \textbf{FPS} \\
				\midrule
				SadTalker & 41.33 & 10.76 \\
				Hallo2    & 45.56 & 0.69  \\
				Sonic     & 25.47 & 0.38  \\
				Float     & 27.19 & 0.47  \\
				Ditto     & 33.99 & 29.85 \\
				\midrule
				\textbf{Ours}
				& \textbf{71.14}
				& \textbf{32.43} \\
				\bottomrule
			\end{tabular}
		}
	\end{minipage}
\vspace{1mm}
\end{figure}

\noindent\textbf{Quantitative Results. }
As shown in Tab.~\ref{tab:compare_all}, TalkLikeYou achieves strong overall performance on both HDTF and CelebV-HQ. It obtains the best Sync-C scores on both datasets, with Sync-D only slightly behind Sonic on HDTF, indicating accurate lip synchronization. Our method also achieves the lowest FID and NIQE, demonstrating better visual quality and naturalness. In addition, TalkLikeYou achieves the best FVD on HDTF and the highest CSIM scores on both datasets, showing strong temporal consistency and identity preservation. As shown in Tab.~\ref{tab:eff}, our method also has an efficiency advantage over Ditto, as it requires only one sampling step during inference. Overall, these results demonstrate that TalkLikeYou achieves balanced performance across lip synchronization, visual quality, motion stability, identity consistency, and efficiency.

\noindent\textbf{Qualitative Results.}
As shown in Fig.~\ref{fig:comparison}, we compare our method with other baselines under three speaking habits. Although these habits involve subtle lip-shape differences for the same pronunciation, our method consistently achieves accurate lip synchronization. In contrast, existing methods often produce unrealistic results: SadTalker produces rigid motions, while Hallo2 and Sonic show blurry teeth regions and unstable lip movements. FLOAT and Ditto yield clearer results, but their motions are less natural, with FLOAT producing limited mouth movements and Ditto often over-opening the mouth. Overall, our method generates high-quality results with fluid motion across different speaking habits. Please refer to the supplementary for user study and more visual results.

\subsection{Evaluation on Habit Imitation}

\noindent\textbf{Similarity.}
We conduct habit imitation experiments against related methods. As shown in Fig.~\ref{fig:adaption}, our method achieves the highest similarity and the lowest JS distance, indicating better imitation of the target speaking habit. The visual clips also show that our results are closest to the reference video, such as in mouth-widening patterns and lower-teeth exposure. More samples are provided supplementary. We also present the histogram of each syllable projection, where our method shows much higher overlap with the reference than other methods, indicating that it more effectively captures the distributional characteristics of the target speaking habit. 

\noindent\textbf{Diversity.}
As shown in Tab.~\ref{tab:eff}, we evaluate the habit diversity of each method. We randomly sample habit features from the habit space and observe that our method achieves a clear advantage over others, indicating that TalkLikeYou can generate more diverse speaking habits rather than collapsing to uniform motions. Furthermore, we observe that each method exhibits characteristic motion patterns, so we can even use motions generated by other methods as references, bridging the motion-distribution gap between each methods. More diversity analysis is provided in the supplementary. 

\subsection{Ablation Study}

\begin{figure}[t]
	\centering
	\begin{minipage}[t]{0.56\linewidth}
		\centering.
		\includegraphics[width=0.95\linewidth]{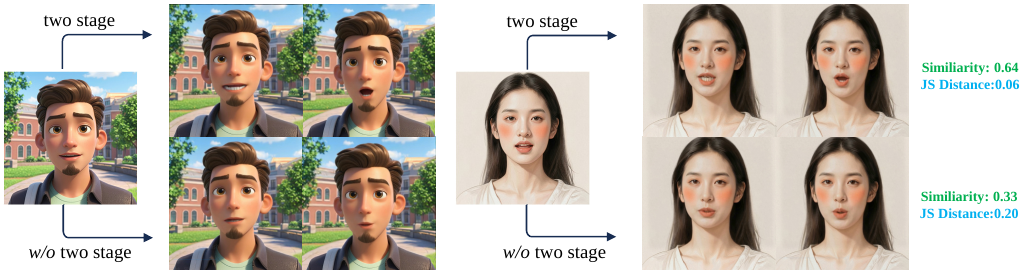}
		\vspace{-2mm}
		\caption{Ablation study on the two-stage habit imitation learning strategy.}
		\label{fig:two_stage}
	\end{minipage}
	\hfill
	\begin{minipage}[t]{0.41\linewidth}
		\centering
		\includegraphics[width=0.9\linewidth]{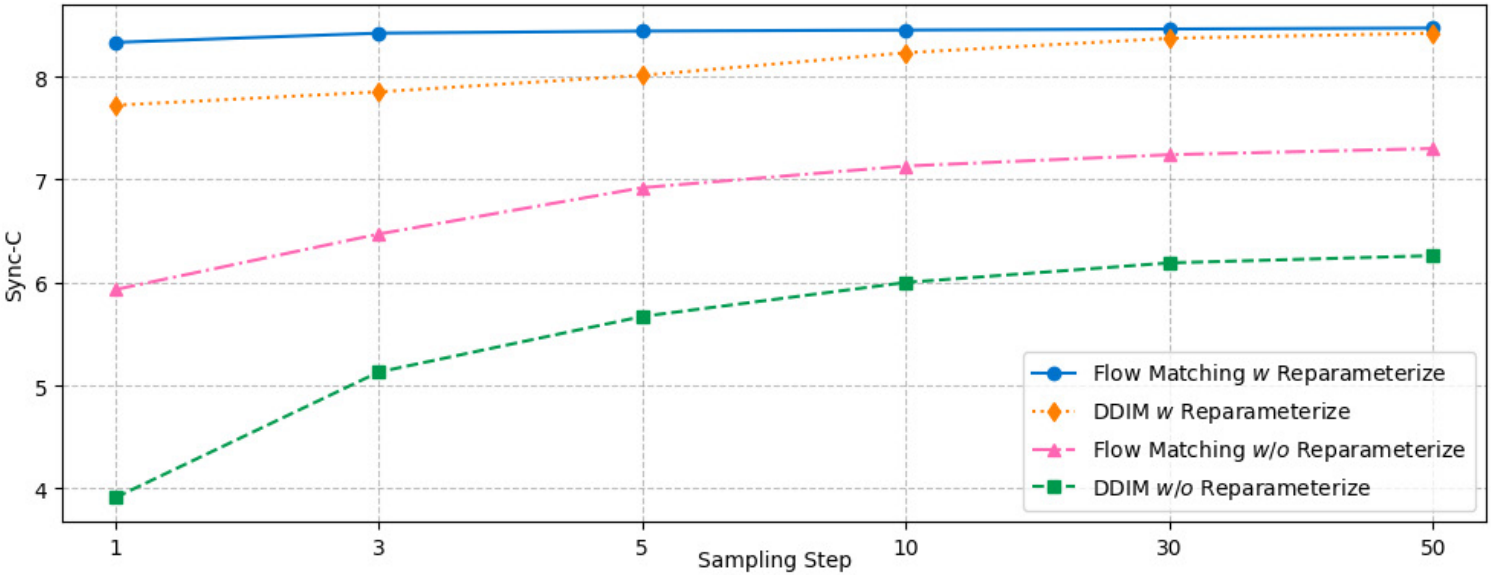}
		\vspace{-2mm}
		\caption{Ablation study of Flow Matching and DDIM with reparameterization.}
		\label{fig:ablation}
	\end{minipage}
\vspace{2mm}
\end{figure}

\noindent\textbf{Two-stage Habit Imitation Learning Strategy.}
As shown in Fig.~\ref{fig:two_stage}, we compare habit imitation results with and without the two-stage learning strategy. The results show that incorporating this strategy improves the generalization of habit motion. In contrast, direct training tends to produce less mouth opening and more uniform motions. The corresponding quantitative metrics further demonstrate the effectiveness of the two-stage learning strategy.

\noindent\textbf{Flow Matching and DDIM with reparameterize. }
We conduct an ablation study on Flow Matching and reparameterization, as shown in Fig.~\ref{fig:ablation}. Flow Matching maintains strong lip-sync performance across sampling steps, while the DDIM-based formulation is less stable, supporting the effectiveness of Flow Matching for efficient motion-space sampling and one-step generation. Removing reparameterization degrades both Flow Matching and DDIM, indicating that conventional velocity or noise prediction leads to unstable generation in our framework. In contrast, reparameterization improves sampling stability and yields clear performance gains. These results validate the importance of both components for accurate and efficient lip-synchronized motion generation.

\section{Conclusion}
In this paper, we present TalkLikeYou, an efficient talking head generation framework for habit imitation. Our method focuses on generating talking head videos with specific speaking habits, providing flexible customization for user requirements. We use Flow Matching to generate motions conditioned on multiple habit sources, enabling efficient one-step sampling during inference. We further adopt a two-stage habit imitation learning strategy to improve habit generalization. To evaluate habit imitation quality, we propose a new metrics PLAD that measures motion similarity and diversity using histogram features from several representative motion axes. Extensive experiments show that our method achieves strong performance in talking head generation and habit imitation, validating the effectiveness of both our framework and the proposed metrics.

\bibliography{arxiv}
\bibliographystyle{iclr2027_conference}

\newpage

\appendix
\section{Appendix}
In the supplementary material, we cover the following aspects:
\begin{enumerate}
	\item \textbf{Further Explanations:} We clarify the speaking-habit imitation task, introduce the Flow Matching formulation, and explain the rationale behind PLAD evaluation.
	\item \textbf{Implementation Details:} We provide further details on the unified habit space, the PLAD algorithm, and the implementation settings.
	\item \textbf{Additional Experimental Results:} We present a user study, an analysis of generation diversity, a visualization of the habit space, and additional qualitative results.
	\item \textbf{Discussion:} We discuss potential directions for future work and ethical considerations for the responsible use of our method.
\end{enumerate}

We strongly encourage readers to watch the supplementary videos to get a clearer sense of our method's motion dynamics and imitation quality. These videos include an overview of our framework, comparisons with existing methods, and demonstrations of speaking-habit imitation. Together, they highlight our method's accurate lip synchronization, faithful habit imitation, and real-time generation capabilities.

\section{Further Explanations}

\subsection{Speaking Habit Imitation}
Speaking-habit imitation aims to generate talking-head videos that reflect how a target person speaks through characteristic facial behaviors, particularly lip movements. Current video generative models can produce high-fidelity videos but often struggle to capture these fine-grained, person-specific articulation patterns, which require focused modeling of the lip region. Therefore, we adopt a motion-based generation method and use implicit keypoints as the representation for their high expressiveness and landmark-based semantic constraints. Specifically, we introduce a landmark-guided loss during implicit keypoint learning to minimize the distance between projected implicit keypoints and 2D facial landmarks. These properties enable us to focus the motion generator on lip-region dynamics, improving speaking-habit imitation.

Several methods have explored speaker-dependent styles related to speaking habits. However, speaking styles and speaking habits describe different levels of behavior. Speaking style broadly characterizes the overall manner of speaking, including rhythm, intonation, emotion, and expressiveness, and can be shared across individuals. In contrast, speaking habits refer to subtle, person-specific articulation patterns that remain relatively consistent across utterances. In this work, we focus on these habitual lip-motion patterns, which capture finer-grained aspects of individual speaking behavior than overall speaking style.

\subsection{Flow Matching} 
\begin{figure}[!t]
	\centering
	\includegraphics[width=\linewidth]{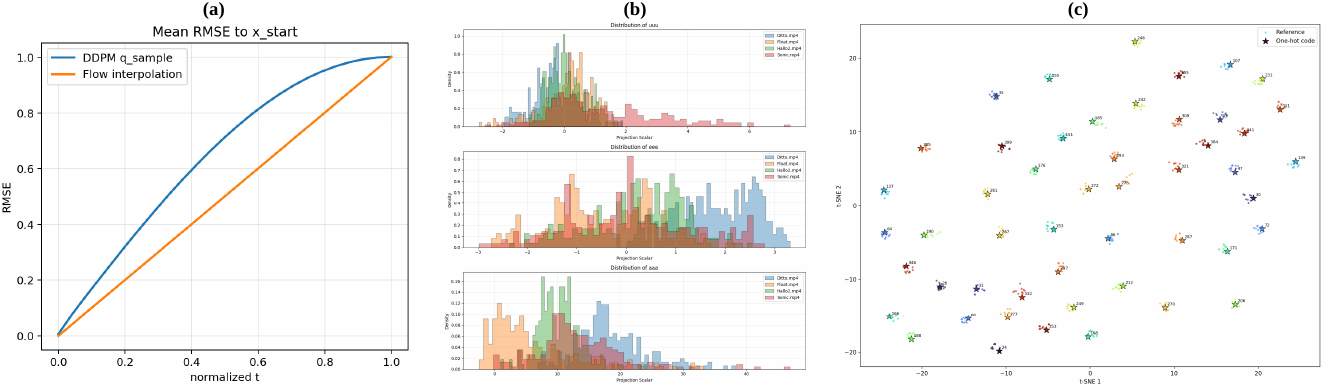}
	\caption{(a) Comparison of generation tendency under DDPM and Flow Matching.
		(b) PLAD histograms of different methods.
		(c) t-SNE visualization of the unified habit space, where stars denote preset one-hot habit features and dots denote reference-based habit features.}
	\label{fig:explain}
	
\end{figure}

Flow Matching is a generative modeling approach that learns a continuous velocity field to transform a noise distribution into a target data distribution. In our framework, it learns to transport noise to clean motion along a linear interpolation path. As shown in Fig.~\ref{fig:explain}(a), the RMSE relative to clean motion changes linearly with the timestep, whereas DDPM exhibits a nonlinear relationship under its noise schedule. This linear interpolation provides a more direct transport path, facilitating approximation of the learned dynamics with fewer sampling steps. In contrast, conventional diffusion sampling typically relies on repeated denoising updates to progressively recover motion from noise. By combining Flow Matching with direct clean-motion prediction, our generator further simplifies inference and produces the complete motion sequence in a single sampling step, supporting efficient, real-time generation.

\subsection{PLAD Evaluation}
Our proposed PLAD uses three representative lip-articulation axes to characterize motion sequences through projection histograms: mouth opening, lip pursing, and lip widening. We consider these axes sufficient to capture key characteristics of speaking habits, as they represent common articulation patterns in daily communication, and many lip configurations can be approximated by their linear combinations. These axes thus provide a compact coordinate system for describing lip motion, analogous to the three spatial coordinates. The histogram along each axis reflects a person's habitual articulation tendencies in the corresponding dimension, collectively characterizing their speaking habit.

Existing motion evaluation protocols rarely assess speaking habits explicitly. In 3D facial animation, metrics such as MSE and RPAC primarily measure reconstruction accuracy and overall motion quality, rather than fine-grained, habit-specific articulation patterns. E-FID~\cite{deng2019accurate} measures distributional differences in expression features across subjects but lacks explicit semantic alignment at the motion level and offers limited insight into specific articulation patterns. In contrast, our PLAD metric projects motion onto three semantically meaningful lip-articulation axes, enabling targeted evaluation of subtle and consistent lip-motion variations that characterize individual speaking habits.

\section{Implementation Details}

\subsection{Unified Habit Space}
One of our goals for habit-aware generation is to support reference-based habit imitation while also providing preset habit control when a reference video is unavailable. To this end, we project the two types of habit representations into a unified habit space. Specifically, we convert the reference video into a motion sequence and use the habit extractor $E_{\mathrm{ref}}$ to obtain its habit representation, while the one-hot habit code is processed by $E_{\mathrm{oh}}$ to produce the preset habit feature:
\[
f_h =
\begin{cases}
	E_{\mathrm{ref}}(M_{\mathrm{ref}}), & \text{reference video is available}, \\
	E_{\mathrm{oh}}(c), & \text{one-hot preset is used}.
\end{cases}
\]

$E_{\mathrm{ref}}$ is based on a Transformer~\cite{vaswani2017attention}, while $E_{\mathrm{oh}}$ projects the one-hot code using an MLP. The habit feature $f_h$ is then fed into the motion generator through both the input layer and the habit-injection layer, together with the diffusion timestep $t$, enabling the generated motion to follow the target speaking habit.

\begin{table}[t!]
	\centering
	\caption{User Study. The rating is on scale of 1-5 where higher rank demonstrate better results.}
		\vspace{-1mm}
	\resizebox{\linewidth}{!}{
		
		\begin{tabular}{c|ccccc}
			\toprule
			\multirow{2}{*}{Methods}  & Lip-sync  & Image  & Temporal & Habit & Overall \\ 
			& Accuracy & Quality & Consistency & Imitation & Realistic\\
			\midrule
			Ditto \cite{li2025ditto} & 2.2 & 2.4 & 2.7 & ----& 2.5 \\
			Hallo2 \cite{cui2025hallo2} & 2.2 & 2.8 & 3.3  & ----& 3.0\\
			Sonic \cite{ji2025sonic}& 4.0 & 2.4  & 3.5  &----& 3.6\\ 
			SadTalker  \cite{zhang2023sadtalker}& 1.2 & 1.9 & 2.4 & ----& 1.9 \\
			StyleTalk  \cite{ma2023styletalk}&1.3 & 1.9 & 1.6 & 1.5& 1.7 \\
			PC-Talk  \cite{wang2025pc}&3.1 & 2.4 & 3.6 & 3.3 &3.4\\ \midrule[1pt]
			\textbf{TalkLikeYou}(Ours)   &\textbf{4.8} & \textbf{4.7} & \textbf{4.6} & \textbf{4.6} &\textbf{4.8} \\
			
			\bottomrule
		\end{tabular}
	}
	\label{tab:user_study}
\end{table}
\begin{figure*}[t]
	\vspace{-4mm}
	\centering
	\includegraphics[width=\textwidth]{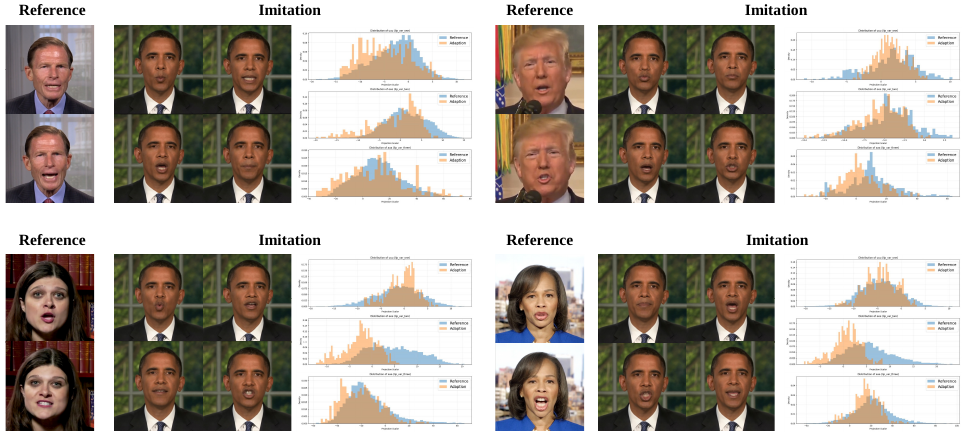}
	\vspace{-5mm}
	\caption{More results on habit imitation.}
	\label{fig:more_habit}
\end{figure*}
\subsection{PLAD Algorithm}
We evaluate speaking-habit similarity and generation diversity separately. For similarity, we first extract lip motions from the generated and reference videos using the motion extractor, then construct three projection histograms for each sequence following the procedure described in the main paper. Each sequence contains at least 500 frames, corresponding to 20 seconds at 25 fps, to obtain stable distribution estimates. We then compute histogram IoU and JS distance between the corresponding generated and reference histograms. We represent each histogram as an array of normalized bin values, allowing direct computation without additional sampling. For diversity, we evaluate more than 50 generated videos and compute the average pairwise JS distance between their PLAD descriptors. This score measures the distributional differences among generated motions in the lip-articulation space, with higher values indicating more diverse speaking habits.

\subsection{Implementation Settings}
We preprocess all videos at 25 fps and extract their audio and motion representations before training. The Flow Matching Motion Generator operates on 18-dimensional lip-motion features and takes a 100-frame reference motion sequence for habit conditioning. The generated motions are then rendered into $512 \times 512$ videos using rendering modules initialized from LivePortrait~\cite{guo2024liveportrait}, which offers a favorable balance between efficiency and visual quality. These rendering modules remain fixed while we optimize the motion generator. The motion extractor is also initialized with LivePortrait. 
We use 1000 flow-matching timesteps and train the generator with Adam~\cite{kingma2014adam} at a learning rate of $1 \times 10^{-4}$. The two training stages take 48 and 72 hours, respectively, on a single RTX 3090 GPU. The first stage uses HDTF~\cite{zhang2021flow}, which contains high-quality videos of 380 identities, each assigned a one-hot code as a preset habit option. The second stage uses CelebV-HQ~\cite{zhu2022celebv}, a large-scale, in-the-wild talking-face dataset, to improve generalization. Throughout training and inference, we use the Wav2Lip~\cite{prajwal2020lip} audio encoder to extract features aligned with lip movements.

At inference time, our framework achieves 32 FPS with one-step Flow Matching sampling. The rendering pipeline animates a cropped face region and stitches it back into the original frame, supporting both image animation and video dubbing. Before animation, we apply a neutralization procedure to establish a consistent initial lip configuration and mouth-opening distance across identities, reducing the effects of imperfect disentanglement in implicit keypoints. The neutral face retains a slightly open mouth to provide a visible reference for the teeth, helping our warp-based renderer preserve tooth structure and maintain temporal consistency throughout the video.

\section{Additional Experimental Results}
\begin{figure*}[t]
	\centering
	\includegraphics[width=\textwidth]{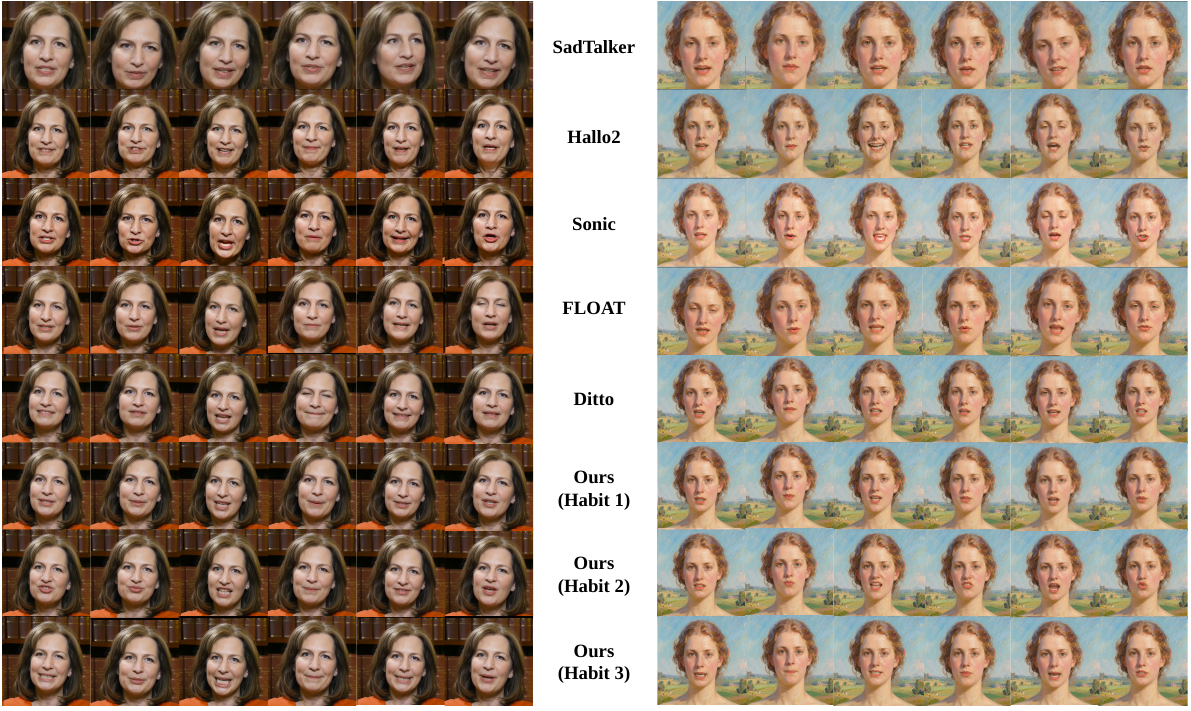}
	\vspace{-4mm}
	\caption{More comparison on Lip Synchronization with other baselines.}
	\label{fig:more_cmp}
\end{figure*}
\subsection{Analysis on Generation Diversity}
We use videos generated by other methods as references to investigate whether different methods exhibit distinguishable motion patterns. As shown in Fig.~\ref{fig:explain}(b), the four methods exhibit distinct distributions in the PLAD space. Sonic tends to produce stronger projections along the lip-pursing axis, which may partly explain its high lip-sync scores if the synchronization metric favors such articulation patterns. Hallo2 tends to generate wider mouth movements, whereas FLOAT and Ditto produce smaller lip movements with more concentrated distributions. These differences suggest that each method implicitly learns characteristic lip-motion patterns, analogous to the distinct speaking habits of individuals. More importantly, our method can use these generated videos as references and adapt to their corresponding motion distributions. This shows that TalkLikeYou is not limited to real-world human reference videos, but can also capture method-specific motion patterns from AI-generated videos. In this sense, our framework provides a flexible platform to bridge the motion-distribution gap across different methods and generate talking motions with controllable, method-like speaking habits.

\subsection{Visualization of the Habit Space}
As shown in Fig.~\ref{fig:explain}(c), we visualize the unified habit space using t-SNE, where stars represent preset one-hot habit features and dots represent features extracted from reference videos. Reference features associated with the same identity cluster around their corresponding preset feature, while different identities form distinct clusters. This suggests that our two-stage learning strategy aligns the two types of habit representations while preserving distinctions among speaking habits, supporting both reference-based imitation and preset habit control within a shared space. Furthermore, by interpolating between two points in the habit space, we can generate a new speaking habit that combines characteristics of both, enabling users to customize speaking habits by blending multiple sources.

\subsection{User Study}
We conduct a user study to evaluate five key dimensions: lip synchronization, image quality, temporal consistency, habit imitation, and overall realism. For each method, we use 10 generated video clips, each lasting more than 5 seconds. The questionnaire follows the Mean Opinion Score protocol with a five-point scale (1--5), where higher scores indicate better performance, and involves more than 20 participants. As shown in Tab.~\ref{tab:user_study}, TalkLikeYou achieves the highest scores across all evaluated dimensions, demonstrating its effectiveness in talking-head generation. In particular, our method performs well in habit imitation, faithfully reproducing target speaking habits while maintaining accurate lip synchronization, high visual quality, and temporal consistency. These results further validate the effectiveness of TalkLikeYou in generating realistic, habit-aware talking-head videos.

\begin{figure*}[t]
	\centering
	\includegraphics[width=\textwidth]{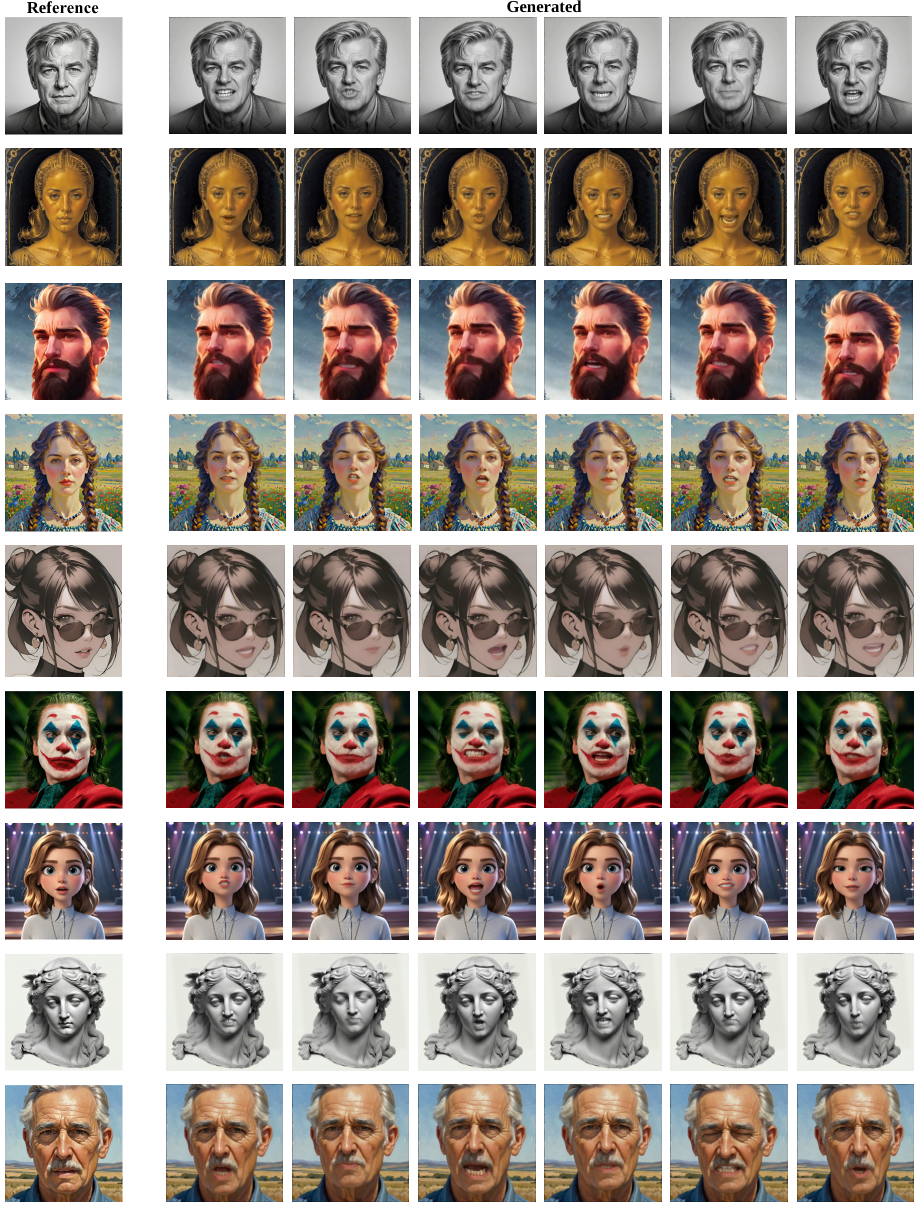}
	\vspace{-5mm}
	\caption{More visualization results on stylize faces.}
	\label{fig:more_stylized}
\end{figure*}
\subsection{More Visible Results}
We provide additional qualitative results for a more comprehensive evaluation. As shown in Fig.~\ref{fig:more_stylized}, our method generalizes to diverse portrait styles, including real photographs, AI-generated images, oil paintings, cartoons, and sketches, demonstrating the robustness of our renderer. We also provide further comparisons with baseline methods in Fig.~\ref{fig:more_cmp}, where our method maintains accurate lip synchronization while supporting diverse speaking habits. Moreover, we present additional habit imitation examples in Fig.~\ref{fig:more_habit}. Our method faithfully imitates the reference speaking habits, with substantial overlap between the generated and reference PLAD histograms, further demonstrating its imitation capability.

\section{Discussion}

\noindent\textbf{Limitations and Future Work.}
While our method can generate talking faces with specific speaking habits, it may struggle to maintain motion stability for exaggerated articulation patterns, such as excessively wide mouth openings. Our framework also focuses primarily on the mouth region, where speaking habits are readily observable, rather than modeling person-specific behaviors across the entire face. In the future, we will explore incorporating eye movements and head pose dynamics to capture a more comprehensive range of individual speaking behaviors.

\noindent\textbf{Ethical Considerations.}
Our method generates high-fidelity talking-face videos that imitate a target person's speaking habits. Since these habits are recognizable aspects of individual behavior, we acknowledge the potential ethical risks and societal concerns associated with misuse, particularly on public platforms. To help mitigate these risks, we are committed to responsible research dissemination, supporting advances in deepfake detection, and promoting the safe and ethical use of generative media technologies.

\end{document}

%% file: math_commands.tex
\usepackage{amsmath,amsfonts,bm}

\def\eqref#1{equation~\ref{#1}}

\def\1{\bm{1}}

\DeclareMathAlphabet{\mathsfit}{\encodingdefault}{\sfdefault}{m}{sl}
\SetMathAlphabet{\mathsfit}{bold}{\encodingdefault}{\sfdefault}{bx}{n}

